\documentclass[conference]{IEEEtran}
\usepackage{cite}
\usepackage{amsmath,amssymb,amsfonts}
\usepackage{graphicx}
\usepackage{textcomp}
\usepackage{xcolor}
\def\BibTeX{{\rm B\kern-.05em{\sc i\kern-.025em b}\kern-.08em
    T\kern-.1667em\lower.7ex\hbox{E}\kern-.125emX}}

\usepackage{graphicx}
\usepackage[utf8]{inputenc} 
\usepackage[T1]{fontenc}    
\usepackage{hyperref}       
\usepackage{url}            
\usepackage{booktabs}       
\usepackage{amsfonts}       
\usepackage{nicefrac}       
\usepackage{microtype}      

\usepackage{tabularx}
\usepackage{color}
\usepackage{algorithm}
\usepackage[noend]{algpseudocode}
\usepackage{multicol}
\usepackage{multirow}

\graphicspath{{images/}}
\newcommand{\lratio}[1]{\setlength{\hsize}{#1\hsize}} 
\newcolumntype{Y}{>{\centering\arraybackslash}X}

\begin{document}

\title{Adversarial Weighting for Domain Adaptation in Regression}

\author{\IEEEauthorblockN{1\textsuperscript{st} Antoine de Mathelin}
\IEEEauthorblockA{\textit{Michelin} \\
Clermont-Ferrand, France \\
antoine.de-mathelin-de-papigny@michelin.com}
\and
\IEEEauthorblockN{2\textsuperscript{nd} Guillaume Richard}
\IEEEauthorblockA{\textit{EDF R\&D} \\
Palaiseau, France \\
guillaume.richard@ens-paris-saclay.fr}
\and
\IEEEauthorblockN{3\textsuperscript{rd} François Deheeger}
\IEEEauthorblockA{\textit{Michelin} \\
Clermont-Ferrand, France \\
francois.deheeger@michelin.com}
\and
\IEEEauthorblockN{4\textsuperscript{th} Mathilde Mougeot}
\IEEEauthorblockA{\textit{Centre Borelli} \\
\textit{Université Paris-Saclay, CNRS, ENS Paris-Saclay}\\
Gif-sur-Yvette, France \\
mathilde.mougeot@ens-paris-saclay.fr}
\and
\IEEEauthorblockN{5\textsuperscript{th} Nicolas Vayatis}
\IEEEauthorblockA{\textit{Centre Borelli} \\
\textit{Université Paris-Saclay, CNRS, ENS Paris-Saclay}\\
Gif-sur-Yvette, France \\
nicolas.vayatis@ens-paris-saclay.fr}
}

\maketitle

\begin{abstract}
    We present a novel instance-based approach to handle regression tasks in the context of supervised domain adaptation under an assumption of covariate shift.
    The approach developed in this paper is based on the assumption that the task on the target domain can be efficiently learned by adequately reweighting the source instances during training phase. We introduce a novel formulation of the optimization objective for domain adaptation which relies on a discrepancy distance characterizing the difference between domains according to a specific task and a class of hypotheses. To solve this problem, we develop an adversarial network algorithm which learns both the source weighting scheme and the task in one feed-forward gradient descent. We provide numerical evidence of the relevance of the method on public data sets for regression domain adaptation through reproducible experiments.
\end{abstract}

\section{Introduction}



Many applications require to learn a regression task, for instance, estimation of manufactured products performance, forecasting of supply and demand or prediction of the time spent by a patient in a hospital. In most of these applications, we observe groups of products or patients defining several domains characterized by different distributions. Acquiring a sufficient amount of labeled data to provide a model performing well on all of these domains is known to be difficult and expensive. In practical cases, no or only a few labeled data are available for the \textit{target} domain of interest whereas a large amount of labeled data are available from another \textit{source} domain. One then seeks to leverage information from the source domain to learn efficiently the task on the target one. This problem is referred as \textit{domain adaptation} \cite{BenDavid2006DATheory}.

The domain adaptation framework is characterized by the presence of a source domain $(Q, f_Q)$ and a target domain $(P, f_P)$ where $Q, P$ define two distributions on the input space and $f_Q, f_P$ are two labeling functions returning the labels for each domain: $y = f_Q(x)$ for any $x \sim Q$ and $y = f_P(x)$ for any $x \sim P$. The main assumption for domain adaptation is that the source and target distributions differ: $P \neq Q$. From there, two common assumptions are considered: the \textit{covariate shift} which states that the source and target domains share the same labeling functions, i.e. $f = f_P = f_Q$ \cite{Bickel2009CovariateShift} and the assumption that the labeling functions match under a specific feature transformation \cite{Courty2016OTDA}. For domain adaptation with a regression task, the covariate shift assumption is often considered \cite{Huang2007KMM, Sugiyama2007KLIEP, Cortes2014DAregression}. In this case, instance-based approaches are mainly used to perform a reweighting of source instances which corrects the shift between input distributions. On the other hand, the second assumption is essentially considered for feature-based approaches which perform a transformation of the input features in order to match the source distribution to the target distribution \cite{Courty2016OTDA}.

In recent years, research in domain adaptation has mainly focused on feature-based approaches due to the success of adversarial neural networks \cite{Ganin2016DANN, Tzeng2017ADDA}. Adversarial domain adaptation methods were originally introduced to learn an encoding space in which the source and target data can not be distinguished by any domain classifier, with the encoder and domain classifier trained with opposing objectives \cite{Ganin2016DANN}. Since then, adversarial training has been used as a powerful tool to minimize complex losses defined as maxima on functional spaces \cite{Shen2017WDANN, Saito2018MCD, Zhang2019MDD, Richard2020MSDA}. The success of adversarial methods lies in their computational speed and ability to scale to large data sets through the use of stochastic gradient descent-ascent algorithms. Moreover, the efficiency of adversarial methods leads to significant improvements for many tasks as image classification \cite{Zhang2019MDD}, image segmentation \cite{Saito2018MCD} or sentiment analysis \cite{Richard2020MSDA}.


However, the developed adversarial methods are exclusively designed to learn a feature transformation through an encoder network. Thus, existing methods are generally not applicable to domain adaptation issues where the covariate shift assumption holds. Indeed, when the source and target labeling functions are the same, the projection of the source instances onto the target instances is likely to impair the learning of the task. We, indeed, observe negative transfer in our numerical experiments when applying feature-based methods on covariate shift domain adaptation issues. On the contrary, instance-based methods often succeed in adapting in this context. However, most of the developed instance-based methods are based on positive semi-definite kernels and their weighting strategy often involves solving a quadratic problem \cite{Huang2007KMM, Cortes2014DAregression, Cortes2019GeneralDisc} which presents a computational burden when the number of data is large.

Our present work aims to address this issue of regression domain adaptation under covariate shift. In this context, the difficulties of negative transfer with unsupervised feature-based methods and the computational burden of instance-based approaches must be overcome. Based on our experience, the scenario where a few labeled target data are available is the most encountered in practice \cite{Cortes2019GeneralDisc}. In this case, supervised methods can greatly improve the performance of domain adaptation \cite{Motiian2017FewShotAdversarial}. These reasons motivate our choice to use a purely supervised metric between source and target distributions: the $\mathcal{Y}$-discrepancy. Furthermore, discrepancy metrics are suited for regression task as they provide theoretical adaptation guarantees on the target risk for common regression losses (as the $L_p$ losses) \cite{Mansour2009DATheory, Medina2015thesis, Cortes2019GeneralDisc, Richard2020MSDA}. We propose an original algorithm called Weighting Adversarial Neural Network (WANN) which relies on an adversarial weighting approach to minimize the $\mathcal{Y}$-discrepancy. The WANN algorithm learns the optimal source weights through a weighting network fitted with adversarial training. Thus WANN takes the best of both worlds by combining the computation efficiency of adversarial neural networks with the ability of instance-based approaches to handle regression domain adaptation under covariate shift.



The organization of the paper is as follows. First, the learning scenario with the definition and the main properties of the $\mathcal{Y}$-discrepancy are introduced (Sections \ref{notation} and \ref{ydisc}). Then, the optimization formulation of the WANN algorithm is presented (Section \ref{optim-form}). A presentation of related works and discussions are given in Section \ref{related}. Finally, numerical experiments are presented on both synthetic and real-world data sets (Section \ref{experiments}).

\section{Weighting Adversarial Neural Network}
\label{wann}

\subsection{Learning scenario}
\label{notation}

Let $\mathcal{X}$ and $\mathcal{Y} \subset \mathbb{R}$ denote the respective input and output spaces. A domain is defined as a pair, composed of a distribution over $\mathcal{X}$ and a labeling function mapping from $\mathcal{X}$ to $\mathcal{Y}$. The source and target domains are respectively written $(Q, f_Q)$ and $(P, f_P)$ with $Q$, $P$ the source and target distributions over $\mathcal{X}$ and $f_Q, f_P: \mathcal{X} \to \mathcal{Y}$ their respective labeling functions.

We consider the supervised domain adaptation scenario inspired from \cite{Cortes2019GeneralDisc}, where the learner has access to a labeled source sample of size $m$, $\mathcal{S} = \{ (x_1, y_1), . . . , (x_m, y_m) \}  \in (\mathcal{X} \times \mathcal{Y})^m$ where $\mathcal{S}_{\mathcal{X}} = \{ x_1, . . . ,x_m \}$ is supposed to be drawn iid according to distribution $Q$ and $y_i = f_Q(x_i)$ for all $i \in [|1, m|]$; and to a labeled target sample of size $n$, $\mathcal{T} = \{ (x_1', y_1'), . . . , (x_n', y_n') \}  \in (\mathcal{X} \times \mathcal{Y})^n$ where $\mathcal{T}_{\mathcal{X}} = \{ x_1', . . . ,x_n' \}$ is assumed drawn iid according to $P$ and $y_i' = f_P(x_i')$ for all $i \in [|1, n|]$. We denote by $\widehat{Q}$ and $\widehat{P}$ the empirical distributions of the respective samples $\mathcal{S}_{\mathcal{X}}$ and $\mathcal{T}_{\mathcal{X}}$. In the supervised domain adaptation setting, $n$ is typically smaller than $m$ ($n << m$). We highlight that, similarly to \cite{Pardoe2010boost}, we do not introduce an unlabeled target sample as the proposed method does not require unlabeled data. We consider in the following that the covariate shift assumption holds, i.e. $f = f_Q = f_P$.

We consider a loss function $L:\mathcal{Y} \times \mathcal{Y} \to \mathbb{R}_{+}$ and a hypothesis set $\mathcal{H}$ of functions mapping from $\mathcal{X}$ to $\mathcal{Y}$. We denote by $\mathcal{L}_D(h, h') = \text{E}_{x \sim D}[L(h(x), h'(x))]$ the average loss (or risk) for any distribution $D$ over $\mathcal{X}$ between two functions $h, h'$. As we consider the instance-based setup where the source losses can be reweighted, we extend the definition of the average loss to functions with a fixed support:  $\mathcal{L}_{\text{q}}(h, h') = \sum_{x \in \mathcal{X}} \text{q}(x) L(h(x), h'(x))$ with $\text{q} \in \mathcal{F}$ and $\mathcal{F}$ a functional space of functions mapping from $\mathcal{X} \to \mathbb{R}_+$.



The learning problem consists in finding a hypothesis $h$ over $\mathcal{H}$ with a small target risk $\mathcal{L}_P(h, f_P)$. 

\subsection{$\mathcal{Y}$-discrepancy}
\label{ydisc}

The $\mathcal{Y}$-discrepancy, first introduced in \cite{Mohri2010Ydiscrepancy}, is defined as the maximal difference between source and target risk over a set of hypotheses. We recall below its definition for any weighting function $\text{q} \in \mathcal{F}$.

\smallskip
\noindent \textbf{Definition 1.}
Let $(Q, f_Q)$ and $(P, f_P)$ be respectively the source and target domains and $\textnormal{q} \in \mathcal{F}$ a weighting function. We define the $\mathcal{Y}$-discrepancy between $P$ and $\text{q}$ as:
\begin{equation}
    \textnormal{$\mathcal{Y}$-disc}_{\mathcal{H}} (P, \textnormal{q}) = \max_{h' \in \mathcal{H}} \, \left| \mathcal{L}_{P}(h', f_P) - \mathcal{L}_{\textnormal{q}}(h', f_Q) \right|
\end{equation}

The $\mathcal{Y}$-discrepancy characterizes how differently a hypothesis from $\mathcal{H}$ can behave between source and target domains. The intuition behind this metric is that the adaptation is successful if the loss obtained on the reweighted sources with any hypothesis $h' \in \mathcal{H}$ is the same as the $h'$ loss obtained on the targets. The $\mathcal{Y}$-discrepancy measures then the worst-case loss gap that a hypothesis $h' \in \mathcal{H}$ can achieve. Notice that, with the covariate shift assumption, the $\mathcal{Y}$-discrepancy is null when $\text{q} = P$.

It can be easily shown \cite{Medina2015thesis} that the $\mathcal{Y}$-discrepancy provides the following generalization bounds of the target risk for any $h\in \mathcal{H}$ and any $\textnormal{q} \in \mathcal{F}$:
\begin{equation}
\label{ydisc-ineq}
    \mathcal{L}_P(h, f_P) \leq \mathcal{L}_{\textnormal{q}}(h, f_Q) + \textnormal{$\mathcal{Y}$-disc}_{\mathcal{H}} (P, \textnormal{q})
\end{equation}

\subsection{Optimization formulation}
\label{optim-form}

From the inequality (\ref{ydisc-ineq}), we infer that finding a reweighting function $\textnormal{q} \in \mathcal{F}$ and a hypothesis $h \in \mathcal{H}$ minimizing $\mathcal{L}_{\textnormal{q}}(h, f_Q) + \textnormal{$\mathcal{Y}$-disc}_{\mathcal{H}} (P, \textnormal{q})$ allows one to learn an efficient hypothesis on the target domain. Since the $\mathcal{Y}$-discrepancy is defined as a maximum on a functional space, we propose to use adversarial training to learn the function $\text{q}$ that minimizes it. To this end, we introduce an adversarial hypothesis $h' \in \mathcal{H}$ to approximate the maximal hypothesis returning the $\mathcal{Y}$-discrepancy. We thus derive this original optimization formulation:
\begin{align}
    \min_{h \in \mathcal{H}, \; \text{q} \in \mathcal{F}} \; \max_{h' \in \mathcal{H}} \;  \mathcal{L}_{\textnormal{q}}(h, f_Q) + \mathcal{L}_{\widehat{P}}(h', f_P) - \mathcal{L}_{\textnormal{q}}(h', f_Q)
\end{align}

The above optimization is a minimization under $h$ and $\text{q}$ of the upper bound from equation (\ref{ydisc-ineq}); except that absolute values were removed because they were not needed to obtain the inequality. Furthermore, by removing them, the optimization objective becomes differentiable and the problem can be solved through stochastic gradient descent-ascent algorithm. Notice that the target loss is computed on the empirical distribution $\widehat{P}$ of the target set $\mathcal{T}_{\mathcal{X}}$. This can be done as a few labels are available on the target domain. Also, similar to \cite{Cortes2019GeneralDisc}, the reweighting function $\text{q}$ is applied on the training set of size $m + n$ obtained by combining $\mathcal{S}$ and $\mathcal{T}$.


Neural networks are used as base hypotheses for the functional spaces $\mathcal{H}$ and $\mathcal{F}$ because of their ability to learn non-linear functions and their fast optimization with stochastic gradient algorithms. A clipping weight regularization, parameterized by the constants $C_h, C_q$, is added to the networks from $\mathcal{H}$ and $\mathcal{F}$ respectively. By using a regularized neural network as weighting function $\text{q} \in \mathcal{F}$, instead of optimizing individual weights for each training instances, the smoothness of the weighting map can be controlled (by the clipping constant $C_q$). With low value of $C_q$, two nearby training points in $\mathcal{X}$ will tend to have similar weights. The algorithm then avoids degenerate weighting schemes where all but a few of the training weights become zero. Adequate value of $C_q$ can be selected through cross-validation on the few available labeled target data.


The architecture of the neural networks from $\mathcal{H}$ as well as the clipping parameter $C_h$ are chosen based on the performances of the hypotheses on the source data. As the covariate shift assumption holds, a good architecture for learning the labeling function on the source domain will likely be adequate for learning this same function on the target domain. The architecture of $\text{q} \in \mathcal{F}$ could be optimized by cross-validation on the few target data. In practice, we set $\mathcal{F} = \mathcal{H}$ for simplicity. However, a ReLU activation is added at end of $\text{q} \in \mathcal{F}$ to ensure positive weights. In addition, to initialize all training weights to $1 / (m+n)$, the network $\text{q}$ is first fitted on training data with $1 / (m+n)$ as output labels.

\section{Related work and Discussion}
\label{related}

\subsection{Adversarial domain adaptation}


Adversarial training of neural networks have been first used for domain adaptation with DANN \cite{Ganin2016DANN}. This method focuses on finding a new representation of the input features where source and target instances cannot be distinguished by any domain classifier. This process aims to minimize the $\mathcal{H}$-divergence \cite{BenDavid2006DATheory}. Since this seminal work, many domain adaptation methods based on adversarial networks have been proposed with other metrics as the Wasserstein's \cite{Shen2017WDANN}, the discrepancy \cite{Saito2018MCD}, the disparity discrepancy \cite{Zhang2019MDD} and the $h$-discrepancy \cite{Richard2020MSDA}. The latter work highlights the advantage of using discrepancy to handle regression tasks over the $\mathcal{H}$-divergence.


The previously mentioned methods consider the unsupervised domain adaptation setting and use then unsupervised distances between distributions. Even though labeled target data can be added to the source data set as \cite{Ganin2016DANN} suggests, some adversarial method focus specifically on the use of labeled samples from the target domain: \cite{Motiian2017FewShotAdversarial, Saito2019SSDA}. These methods, however, focus only on classification tasks and can hardly be extended to regression.


\subsection{Adversarial domain adaptation under target shift}

Previous works on adversarial domain adaptation mainly focus on feature-based approaches. However, recent works introduce importance weighting in adversarial methods especially for partial domain adaptation (PDA) which considers the target shift issue. Target shift occurs when the output distribution differ between the source and target domain \cite{Zhang2013DAUnderTargetShift}. Some PDA methods propose to reweight the source instances during the training process to "discard" the ones unrelated to the target domain. For instance, \cite{Chen2018ReweightedAdversarialAdaptation, Li2019TargetShiftAdversarialDA} propose for classification task, to give an importance weight to each class in the source domain. Other approaches propose to weight specifically each source instance \cite{Zhang2018ImportanceWeightedAdversarial4PDA, Cao2019LearningToTransfer4PDA}, the proposed weighting scheme is obtained by using the output of one domain classifier. Notice that these methods learn at the end a new representation of the input features with adversarial training.

A work closely related to ours \cite{Binkowski2019batchWeightingDA} explicitly introduces a weighting network to reweight batches during a GAN training.
However, their framework, in line with cycle-GAN, differs from ours as they focus on the divergence given by a domain discriminator ($\mathcal{H}$-divergence) instead of the $\mathcal{Y}$-discrepancy.

\subsection{Instance-based domain adaptation and curriculum learning}

Instance-based domain adaptation methods propose to correct the difference between source and target distributions by reweighting the source instances. Most of them propose to minimize a specific distance between distributions as for instance the KL divergence \cite{Sugiyama2007KLIEP}, the MMD \cite{Huang2007KMM}, the discrepancy and other related distances \cite{Mohri2010Ydiscrepancy, Cortes2014DAregression, Mansour2009DATheory, Cortes2019GeneralDisc}. Other methods consider boosting iterations as \cite{Pardoe2010boost} for example. Most of these methods consider positive semi-definite kernels and propose to solve a quadratic problem \cite{Cortes2014DAregression, Cortes2019GeneralDisc, Huang2007KMM} which hardly scales to large samples.

A work on Bayesian instance-based transfer learning \cite{Ruder2017BayesianInstanceBased} highlights the links between instance-based domain adaptation and curriculum learning \cite{Bengio2009CurriculumLearning}. Indeed curriculum learning methods propose to reweight the training instances. Some of these methods have also been used for domain adaptation \cite{Zhang2017CurriculumDA, Zou2018UnsupervisedDACurriculum}. However, as mentioned in \cite{Ruder2017BayesianInstanceBased}, curriculum learning aims at "ordering" the training data whereas instance-based domain adaptation aims at "selecting" the best source sample for adaptation.

\subsection{Discussion}
\label{discuss}

Most domain adaptation methods focus on the feature-based framework, where the goal is to learn a transformation of the feature space $\phi$ correcting the domain shift between $Q$ and $P$. The assumption commonly made in this framework is that the labeling functions of the two domains are matching after the transformation, i.e. $f_P(x) = f_Q(\phi(x)) \; \forall \, x \in \mathcal{X}$ in the asymmetric setting \cite{Courty2016OTDA}, \cite{Tzeng2017ADDA}, or $f_P(\phi(x)) = f_Q(\phi(x)) \; \forall \, x \in \mathcal{X}$ in the symmetric one \cite{Ganin2016DANN}, \cite{Shen2017WDANN}. This assumption is usually motivated by the existence of domain-invariant features \cite{Ganin2016DANN} or by a change in the acquisition conditions between the two domains \cite{Courty2016OTDA} (noise, sensor drift...). The covariate shift assumption states that the labeling functions match for $\phi = \text{Id}$. Applying, in this context, another transformation $\phi$ as for instance, optimal transport \cite{Courty2016OTDA} or encoding networks trained to minimized the $\mathcal{H}$-divergence \cite{Ganin2016DANN}, will probably produce negative transfer.

\begin{figure}[ht]
	\begin{center}
		\includegraphics[width=\columnwidth]{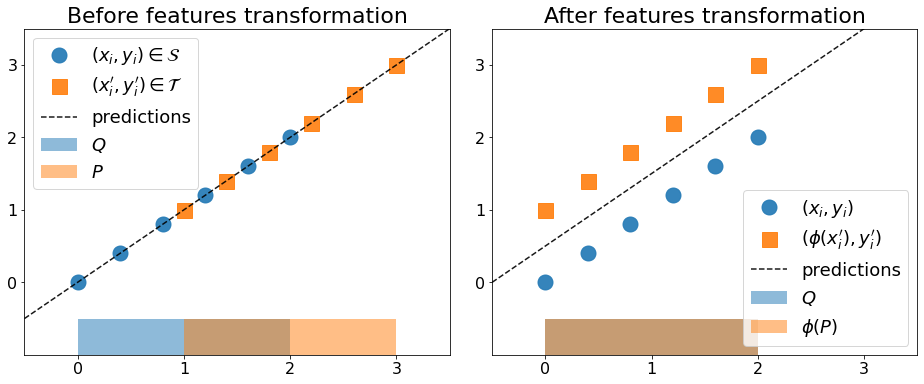}
		\caption{Negative transfer caused by feature-based methods. Matching $Q$ and $P$ is detrimental here, as the shift between the two distributions is informative with respect to the task.}
		\label{illustrative_example}
	\end{center}
\end{figure}

In an industrial design scenario for instance, the learner aims at building an accurate predictive model for new products based on an historical labeled data set from previous products. In this kind of regression problems, the shift between the input distribution $P$ and $Q$ is likely to be informative and looking for a perfect matching of the two distributions has no sense. For a better understanding of this problem, we propose the following illustrative example: considering $Q$ as a uniform distribution between $[0, 2]$, i.e $Q = \mathcal{U}[0, 2]$ and $P = \mathcal{U}[1, 3]$ and $f_Q = f_P = \text{Id}$, in this case, matching $Q$ and $P$ with optimal transport \cite{Courty2016OTDA} or with domain adversarial neural networks \cite{Tzeng2017ADDA}, will lead to negative transfer as it is observed in Figure \ref{illustrative_example}. Notice that, in this illustrative example, an instance-based method will not perform better than a standard approach but will avoid negative transfer.

\section{Experiments}
\label{experiments}

In this section, we report the results of WANN algorithm
on several experiments. The first set of experiments is based on a synthetic data set proposed by \cite{Cortes2014DAregression} on which we compare for several input dimensions the WANN algorithm to the naive approaches: \textbf{Uniform Weighting} and \textbf{Target Only}.

The second set of experiments are conducted on two real-world public data sets: CityCam \cite{Zhang2017WebCamT} and Amazon reviews \cite{Blitzer2007SA} which consist respectively in a counting task on images and a sentiment analysis on text. The WANN algorithm is compared to the most representative domain adaptation methods from both feature-based and instance-based strategies. The selected feature-based methods are the following:

\begin{itemize}
    \item \textbf{DANN} \cite{Ganin2016DANN} is used here for regression tasks by considering the mean squared error as task loss instead of the binary cross-entropy proposed in the original algorithm. In the following DANN uses all labeled data to learn the task and all available training target data (including unlabeled ones) to train the domain classifier.
    \item \textbf{ADDA} \cite{Tzeng2017ADDA} performs a DANN algorithm in two-stage: it first learns a source encoder and a task hypothesis using labeled data and then learns the target encoder with adversarial training. 
    \item \textbf{Deep-CORAL} \cite{Sun2016CORAL} learns a new feature representation by minimizing the difference between the correlation matrix of encoded source and target data.
    \item \textbf{MDD} \cite{Zhang2019MDD} learns a new feature representation by minimizing the disparity discrepancy between the encoded source and target domains.
\end{itemize}

\noindent The selected instance-based competitors are the following:

\begin{itemize}
    \item \textbf{Uniform Weighting} assigns a uniform weight to every instance.
    \item \textbf{Target Only} uses only labeled target data.
	\item \textbf{TrAdaBoostR2} \cite{Pardoe2010boost} is based on a reverse-boosting principle where the weight of source instances poorly predicted are decreased at each boosting iteration. $10$ boosting iterations are used.
	\item \textbf{KLIEP} \cite{Sugiyama2007KLIEP} is a kernel-based sample bias correction method minimizing the KL-divergence between a reweighted source and target distributions. The default number of maximal centers: $100$ is used.
	\item \textbf{KMM} \cite{Huang2007KMM} reweights source instances in order to minimize the MMD between domains. A	gaussian kernel and the default parameters $B=1000$ and $\epsilon = (\sqrt{m}-1) / \sqrt{m}$ are used.
\end{itemize}

The source code of the selected domain adaptation methods is provided in the python library ADAPT\footnote{\url{https://github.com/adapt-python/adapt}} \cite{Demathelin2021ADAPT}, the source code of WANN and the scripts used to obtain the presented results are available on GitHub\footnote{\url{https://github.com/anonymousaccount0/wann}}. The presented results of this section are computed on a ($2.7$ GHz, $16$ G RAM) computer. For all neural network optimizations, the Tensorflow library is used \cite{tensorflow2015whitepaper} as well as the Adam optimizer \cite{Kingma2014Adam}. If nothing else is specified, the optimization parameters used in the presented experiments are $\textit{lr} = 0.001$, $\beta_1=0.9$ and $\beta_2=0.999$ and the loss function is the mean squared error (MSE).


\subsection{Synthetic data set}

The first set of experiments aim to test the efficiency of the WANN algorithm in a controlled environment. We consider the regression domain adaptation issue proposed by \cite{Cortes2014DAregression} which satisfies the covariate shift assumption (the source and target domains share the same labeling function). In the proposed setup, the source input distribution $Q$ is a mixture of $N$ gaussians with bandwidth equal to $1$ and centered in the hyper-cube $[-1, 1]^N$. The target distribution is a single gaussian also with bandwidth equal to $1$ and centered in $[-1, 1]^N$. The source distribution contains $20 \%$ of instances drawn according to the target gaussian distribution, these instances are then considered as target labeled instances in the WANN algorithm. The labeling function $f$ is defined as the mean of absolute values of the input vector's components and is shared by the two domains: $f(x) = \frac{1}{N} \sum_1^N |x_i|$ with $x = (x_1, ..., x_N) \in \mathbb{R}^N$. The base hypothesis used to learn the task is a neural network with two hidden fully-connected layers of $100$ neurons each, ReLU activation functions and weights clipping $C_h = 1$; $300$ epochs with a batch size of $128$ are performed.

\begin{figure*}[ht]
	\begin{center}
		\includegraphics[width=\textwidth]{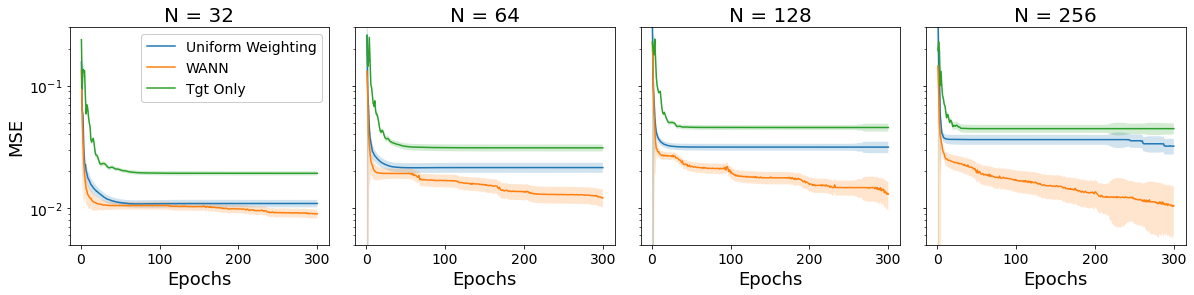}
		\caption{Synthetic experiments results: target MSE evolution across epochs for four values of $N$.}
		\label{scores_synth}
	\end{center}
\end{figure*}

We consider a number of source instance $m = 1000$ from which around $200$ instances are drawn according to the target distribution, we also consider $1000$ target instances used as validation data. Following \cite{Cortes2014DAregression}, we vary the dimension $N$ of the input space $N \in \{32, 64, 128, 246 \}$. We report on Figure \ref{scores_synth} the MSE evolution computed on the target validation data through the $300$ epochs for the three algorithms: WANN, Uniform Weighting and Target Only. The results are averaged on $10$ runs to provide confidence intervals (in light color).

\begin{figure}[ht]
	\begin{center}
		\includegraphics[width=7cm]{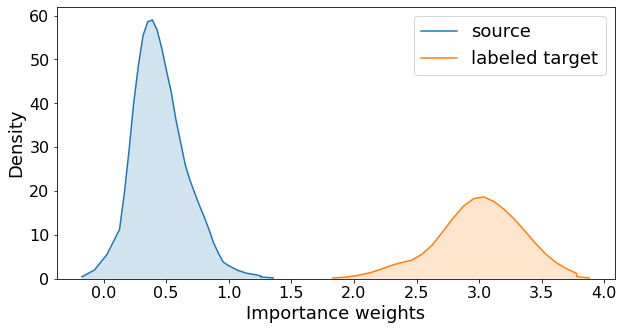}
		\caption{Distribution of training weights returned by the WANN weighting network for $N=256$. The weights are normalized so that the mean is equal to one.}
		\label{weights_synth}
	\end{center}
\end{figure}

We observe that the difficulty of the task is increasing with the number of dimension $N$ as the target MSE of Uniform Weighting and Target Only increase with the dimension. It appears clearly that WANN outperforms the two other approaches, especially when the task difficulty increases. Indeed, we observe that WANN MSEs on target data decrease faster and lower than the MSEs of the two naive methods. These results highlight the efficiency of WANN to handle regression domain adaptation issue under covariate shift. The distribution of importance weights returned by the weighting network of the WANN algorithm (cf Figure \ref{weights_synth}) presents two distinct modes corresponding to the $20 \%$ of source instances drawn according to the target gaussian distribution (in orange) and the $80 \%$ other source instances (in blue). We observe that the weighting network assigns higher weights to the first group of instances (around $3$) while keeping a reasonable weighting on other instances (around $0.5$).

\subsection{Real-world data sets}

For the experiments with real-world data sets, we compare the results obtained with the WANN algorithm to the ones obtained with the previously described competitors. To compare only the adaptation effect of each method, we use, for all of them, the same class of hypotheses $\mathcal{H}$ to learn the task which is the class of fully-connected neural networks with ReLU activation functions and a static architecture. All networks from $\mathcal{H}$ implement a weight clipping regularization of parameter $C_h$. For WANN algorithm, the two networks $h$, $h'$  are chosen in the specified class $\mathcal{H}$. The weighting network $\text{q}$ is taken in the functional space $\mathcal{F}$, if nothing is specified we choose $\mathcal{F}=\mathcal{H}$ with the same clipping parameter $C_{\text{q}} = C_{h}$. Notice, however, that we add a ReLU activation function at the end of $\text{q}$ to ensure positive source weights. For feature-based methods, we split the networks $h \in \mathcal{H}$ into an encoder part $\phi$ and a task part $g$ such that $h = g \circ \phi$. Thus the learning hypothesis for the task has the same complexity for each method.

The present work focuses on the covariate shift domain adaptation setting in which the source and target labeling functions match. To test whether this assumption holds for the selected real-world data sets we conduct the following experiment: we compute cross-validation scores on the source and target domain for different training data sets: those composed of the respective instances of each domain and those containing the instances of both domains. We assume that if the source and target cross-validation scores obtained with models trained on both domain are as good as those obtained with models trained on single domains, it means that the source and target labeling functions are close. The obtained cross-validation scores for CityCam are reported in Table \ref{covariate-test}. We observe that the scores are close and therefore consider the covariate shift to be a reasonable assumption for these experiments. A similar observation is made for experiments on Amazon reviews. 


\begin{table}[ht]
	\small
	\centering
	\caption{CityCam cross-validation scores for testing covariate shift.}
	\label{covariate-test}
	\begin{tabularx}{\columnwidth}{>{\lratio{0.85}}X>{\lratio{1.05}}X>{\lratio{1.05}}X>{\lratio{1.05}}X}
		\toprule
Train set		&  Source  & Target    & Both    \\
        \midrule
Source  & 1.58 (0.04) & /  &  1.59 (0.02) \\
Target  & /  & 1.39 (0.08) & 1.39 (0.04) \\
\bottomrule
	\end{tabularx}
\end{table}

\subsubsection{CityCam vehicle counting data set}
\label{citycam-exp}

For the first real-world experiment, we consider the CityCam vehicle counting data set \cite{Zhang2017WebCamT}. This data set is composed of images of city camera videos coming from several cameras distributed in different city locations. The data set is annotated, including for each image the box location of each vehicle and the total number of vehicles. We focus on predicting the number of vehicles. This task is useful, for instance, for understanding the traffic density or for detecting traffic jams. This data set has been previously used for regression domain adaptation in the work of \cite{Zhao2018MultiSourceDANN} considering the images from each camera as belonging to separated domains. To demonstrate the effectiveness of WANN in this setting, we select images coming from $4$ cameras: two located on a highway and the two others located at an intersection. The images from one of the intersection camera are selected as target data ($\sim 5000$ data) and the images from the three other cameras as source data ($\sim 10 000$ data). We use, as input features, the outputs from ResNet50 \cite{He2016ResNet} to which we further apply a standard scaling using the source input data. We also rescale the counting number of vehicles using the source output labels. We select the best trade-off parameter $\lambda$ for DANN, Deep-CORAL and MDD and the best kernel bandwidth for KMM and KLIEP among several tested ones.






We conduct each experiment 10 times and report the mean and standard deviation of mean absolute errors (MAEs) for each method in Table \ref{citycam-results}. We randomly pick $n$ target data as labeled target data set. We vary $n \in \{20, 50, 100, 200 \}$. The other target instances are used as unlabeled target data by feature-based methods and KMM and KLIEP; they are also used to compute the MAEs.

\begin{table}[ht]
	\small
	\centering
	\caption{MAE between the true vehicle count and the
            estimated count on the CityCam data set.}
	\label{citycam-results}
	\begin{tabularx}{\columnwidth}{>{\lratio{0.6}}X>{\lratio{1.1}}X>{\lratio{1.1}}X>{\lratio{1.1}}X>{\lratio{1.1}}X}
		\toprule
$n$		& 20          & 50          & 100         & 200  \\
        \midrule
Tgt O.      & 3.05 (0.18) & 2.69 (0.10)  & 2.43 (0.07) & 2.16 (0.11) \\
Unif.   & 3.20 (0.23)  & 2.91 (0.16) & 2.63 (0.10)  & 2.24 (0.06) \\
\midrule
DANN         & 4.92 (1.25) & 4.01 (1.59) & 3.26 (0.53) & 2.71 (0.21) \\
ADDA         & 3.21 (0.17) & 3.78 (0.79) & 3.75 (1.24) & 3.71 (0.98) \\
CORAL    & 5.37 (0.27) & 4.52 (0.50)  & 3.86 (0.41) & 2.87 (0.21) \\
MDD          & 3.35 (0.41) & 3.03 (0.28) & 2.78 (0.23) & 2.42 (0.15) \\
\midrule
TrAdaB. & 3.30 (0.18)  & 2.96 (0.13) & 2.52 (0.16) & 2.28 (0.11) \\
KLIEP        & 3.05 (0.48) & 2.60 (0.15)  & 2.34 (0.1)  & 2.07 (0.07) \\
KMM          & \textbf{2.79 (0.16)} & 2.52 (0.10)  & 2.32 (0.07) & 2.06 (0.09) \\
WANN         & \textbf{2.79 (0.22)} & \textbf{2.48 (0.08)} & \textbf{2.26 (0.04)} & \textbf{1.98 (0.07)} \\
		\bottomrule
	\end{tabularx}
\end{table}

\begin{figure*}[ht]
	\begin{center}
		\includegraphics[width=\textwidth]{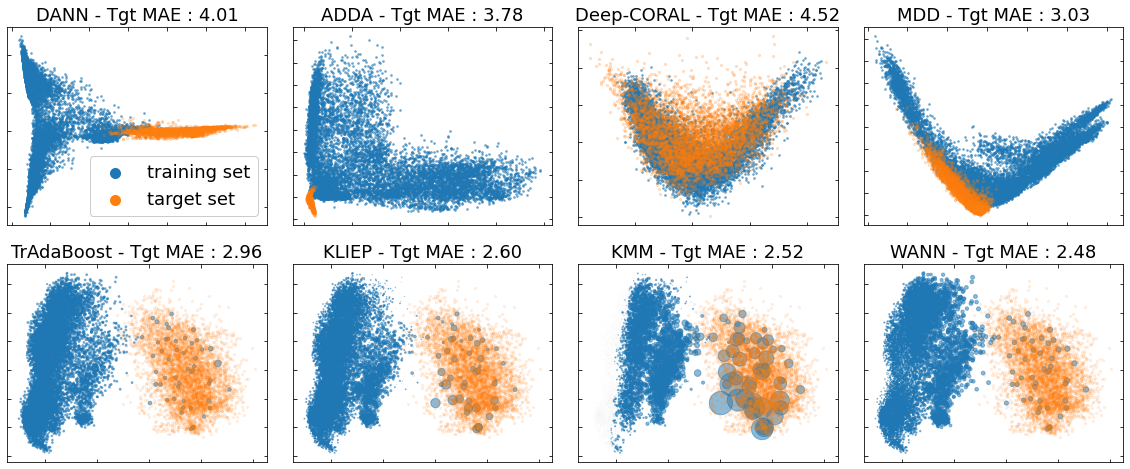}
		\caption{Two first PCA components of the encoded space for the four plots at the top and of the input space for the four plots at the bottom. For instance-based methods, the sizes of the points are proportional to the importance weights.}
		\label{pca_citycam}
	\end{center}
\end{figure*}





We observe on Table \ref{citycam-results} that WANN, KMM and KLIEP provide the best performances, which underlines that instance-based approaches are more suited than feature-based ones for adaptation under the covariate shift assumption. In fact, none of the feature-based approaches provide a better performance than the two naive strategies: Uniform Weighting (Unif.) and Target Only (Tgt O.) and thus perform negative transfer. We also observe that the gap between WANN results and the ones of KMM and KLIEP progressively increases when the number of labeled target data in the training set increases. This can be explained by the fact that WANN, in the contrary of KMM and KLIEP, is based on the $\mathcal{Y}$-discrepancy which is a purely supervised domain adaptation metric, thus the WANN performance improves as the estimation of $\mathcal{L}_{\widehat{P}}(h', f_P)$ become more accurate.

\begin{figure}[ht]
	\begin{center}
		\includegraphics[width=\columnwidth]{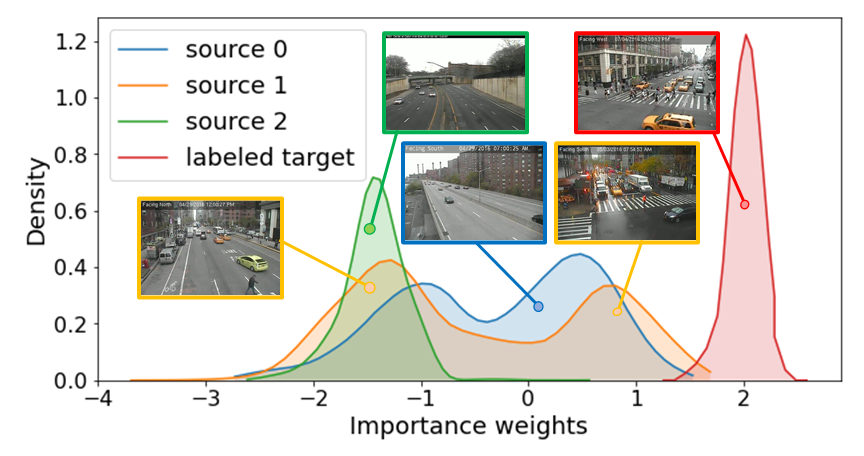}
		\caption{Distribution of training weights returned by the weighting network of WANN in the CityCam experiment with $n=50$. The weights are normalized and log scaled for a better visualization.}
		\label{weights_citycam}
	\end{center}
\end{figure}

We report on Figure \ref{pca_citycam} the two first components of the PCA in the encoded space for the feature-based methods and in the input space for instance-based methods. We observe that feature-based methods manage to make source and target data look close in the encoded space. However, this is detrimental for their performances on target data. We observe indeed that Deep-CORAL performs the best match between source and target distribution but also presents the higher target MAE. On the contrary, instance-based methods provide reasonable target MAEs. We observe that these methods assign higher weights to the target labeled data and to source instances close from the target domain. We also observe that the weighting scheme of WANN is more conservative than the one of KMM and KLIEP. Figure \ref{weights_citycam} provides the detail of the WANN importance weights according to each source, we observe that the instances of the source intersection camera (source $1$) have in average higher weights than other source instances. It is interesting to notice that source $1$ importance weights present two modes because the images from this source come from two different views, the higher mode corresponding to the intersection view. This highlights the ability of WANN algorithm to catch useful source instances for the learning on the target domain.




\subsubsection{Amazon reviews sentiment analysis}
\label{comparison-instance}

We conduct the second experiment on the cross-domain sentiment analysis data set of Amazon reviews \cite{Blitzer2007SA} where reviews from four domains: dvd, kitchen, electronics and books are rated in $[|1, 5|]$. The task consists to predict the rating given one review. The data set has been used for regression domain adaptation experiments in \cite{Cortes2019GeneralDisc} considering one domain as source and another as target, defining then $12$ experiments. We follow the settings from \cite{Cortes2019GeneralDisc} using the top $1000$ uni-grams and bi-grams as input features, $700$ randomly picked labeled source and unlabeled target data and $50$ labeled target data. The results are computed on $1000$ target data. Each of the $12$ experiment is conducted $10$ times to get standard deviation indices. The base network $h \in \mathcal{H}$ used to learn the task is composed of $2$ layers of respectively $100$ and $10$ neurons with parameter $C_h = 1$. As we encounter over-fitting when training the network on source instances, dropouts is added at the end of each layer with rates $(0.8, 0.5)$. These rates are chosen based on validation scores computed on the source domain. We use cross-validation on the $50$ labeled target data to set hyper-parameters $C_q$ for WANN, $\lambda$ for DANN, Deep-CORAL and MDD and $\sigma$ for KMM and KLIEP. $200$ epochs are performed with a batch size of $128$.

\begin{table*}[ht]
\caption{Amazon reviews sentiment analysis results summary (MSE $\times 1000$). Averages are done over the $12$ experiments.}
\label{results-sa}
\centering
\begin{tabularx}{\textwidth}{X>{\lratio{0.9}}YYYYYYYYY>{\lratio{1.1}}Y}
\toprule
  & Tgt O.   & Unif. & WANN     & TrAdaB. &  KMM  & KLIEP & DANN & ADDA  & CORAL & MDD \\
  \midrule
Avg MSE & 1066 (41) & 999 (39)   & 988 (40) & 984 (38)     & 1031 (48) & 1059 (64) & 1071 (61) & 1116 (52) & 1008 (44) & 1126 (153) \\
Median & 1068 (38) & 1009 (40)  & 995 (43) & 1003 (34)    & 1039 (47) & 1068 (65) & 1064 (63) & 1118 (47) & 1012 (47) & 1109 (121) \\
Avg rank & 6.83      & 3.33       & 1.83     & 2.0          & 5.0       & 6.92      & 7.17      & 9.0       & 4.08      & 8.83       \\
$1^{\text{st}}$ rank & 0         & 0          & 7        & 5            & 0         & 0         & 0         & 0         & 0         & 0          \\
Top $3$ & 2         & 7          & 11       & 10           & 2         & 0         & 0         & 0         & 4         & 0        \\
\bottomrule
\end{tabularx}
\end{table*}

A summary of the results is reported on Table \ref{results-sa}. For these experiments, it appears that WANN provides the more interesting performances although it should be notice that the advantage of using WANN is less significant than for previous experiments. We compute, however, the probabilities that one algorithm is better than the others according to an extension for regression task of the pairwise Poisson binomial test \cite{Ganin2016DANN}. We observe that WANN and TrAdaBoost are better than other algorithms with probability $0.67$ against Unif, $0.74$ against Deep-CORAL and more than $0.9$ against the others. WANN and TrAdaBoost provide similar performances, but TrAdaBoost is an ensemble method which takes advantage of averaging the predictions over several models. To make a fair comparison, we conduct the same experiments with a bagging of WANN models and observe that WANN has then a significantly better performance than TrAdaBoost with probability $0.62$.



\section{Conclusion}

This work presents a novel instance-based approach for regression tasks in the context of supervised domain adaptation under the covariate shift assumption. We show that the training weights can be optimally adjusted with a neural network in order to efficiently learn the target task. This is achieved by the WANN algorithm which provides, on various experiments, results which outperform the domain adaptation baselines. We show that feature-based methods are not suited to handle regression domain adaptation when the covariate shift assumption holds. Our work also highlights the efficiency of using a supervised metric: the $\mathcal{Y}$-discrepancy when a few labeled target data are available. Future work will focus on the extension of WANN to the unsupervised and semi-supervised domain adaptation framework as well as the ability of the weighting network to provide useful information on the next target labels to query from an active learning perspective.

\bibliographystyle{plain}
\typeout{}
\bibliography{references}

\end{document}